\documentclass{article}

\PassOptionsToPackage{numbers, compress}{natbib}

\usepackage[preprint]{neurips_2025}

\usepackage[utf8]{inputenc} 
\usepackage[T1]{fontenc}    
\usepackage{hyperref}       
\usepackage{url}            
\usepackage{booktabs}       
\usepackage{amsfonts}       
\usepackage{nicefrac}       
\usepackage{microtype}      
\usepackage[table,xcdraw]{xcolor}         
\usepackage{multirow}
\usepackage{graphicx}
\usepackage{caption}  
\usepackage{amsmath, amssymb}
\usepackage[ruled,vlined]{algorithm2e}

\title{TopKD: Top-scaled Knowledge Distillation}

\author{%
  Qi Wang\\
  Hosei University\\
  Tokyo, Japan \\
  \texttt{qi.wang.7c@stu.hosei.ac.jp} \\
  \And
  Jinjia Zhou \\
  Hosei University \\
  Tokyo, Japan \\
  \texttt{zhou@hosei.ac.jp} \\
}

\begin{document}
\maketitle
\begin{abstract}
   Recent advances in knowledge distillation (KD) predominantly emphasize feature-level knowledge transfer, frequently overlooking critical information embedded within the teacher's logit distributions. In this paper, we revisit logit-based distillation and reveal an underexplored yet critical element: Top-K knowledge. Motivated by this insight, we propose Top-scaled Knowledge Distillation (TopKD), a simple, efficient, and architecture-agnostic framework that significantly enhances logit-based distillation. TopKD consists of two main components: (1) a Top-K Scaling Module (TSM), which adaptively amplifies the most informative logits, and (2) a Top-K Decoupled Loss (TDL), which offers targeted and effective supervision. Notably, TopKD integrates seamlessly into existing KD methods without introducing extra modules or requiring architectural changes. Extensive experiments on CIFAR-100, ImageNet, STL-10, and Tiny-ImageNet demonstrate that TopKD consistently surpasses state-of-the-art distillation methods. Moreover, our method demonstrates substantial effectiveness when distilling Vision Transformers, underscoring its versatility across diverse network architectures. These findings highlight the significant potential of logits to advance knowledge distillation.
\end{abstract}
\section{Introduction}
\label{introduction}
Knowledge Distillation (KD), introduced by Hinton et al. \cite{kd}, aims to transfer the generalization capabilities of a powerful teacher model to a lightweight student model by aligning their predictions. Since then, KD has been widely adopted due to its practicality, efficiency, and versatility. It has demonstrated applicability across various network architectures and can be seamlessly integrated with other compression techniques, such as pruning \cite{pruning1,pruning2} and quantization \cite{quantization1, quantization2, kdquntify}, to further reduce model size.\par

Despite the extensive success of KD, most state-of-the-art methods \cite{similarity, ab, factor, crd, review, reuse, fcfd} predominantly focus on distilling feature representations from intermediate layers of the teacher model. While feature-based distillation effectively transfers rich semantic representations, it typically incurs high computational costs, complex training procedures, and strong architectural dependencies. In contrast, logit-based distillation \cite{kd, dkd, dot, ctkd} offers a more lightweight and flexible alternative by supervising the student through the teacher’s output distributions, making it easier to deploy across diverse models.\par

However, logit-based distillation methods have generally exhibited inferior performance compared to feature-based distillation methods, primarily due to an information bottleneck at the final output layer where distillation supervision is usually restricted to class probabilities. This limitation is further exacerbated by the widespread use of Kullback-Leibler divergence (KL-Div) since \cite{kd}, which enforces a strict alignment between student and teacher logits. While effective in certain settings, this rigid alignment often restricts the richness of knowledge transfer and can hinder student generalization. Recognizing this limitation, recent studies \cite{dkd, dot, ttm, wasserstein} have explored alternative formulations by either relaxing or entirely removing the KL-Div constraint, demonstrating that rethinking logit-based supervision can yield competitive or superior results. These developments highlight the untapped potential of logit-based distillation, motivating a deeper investigation into its structure and learning dynamics.\par
\begin{figure}[htbp]
	\centering
	\includegraphics[width=\textwidth]{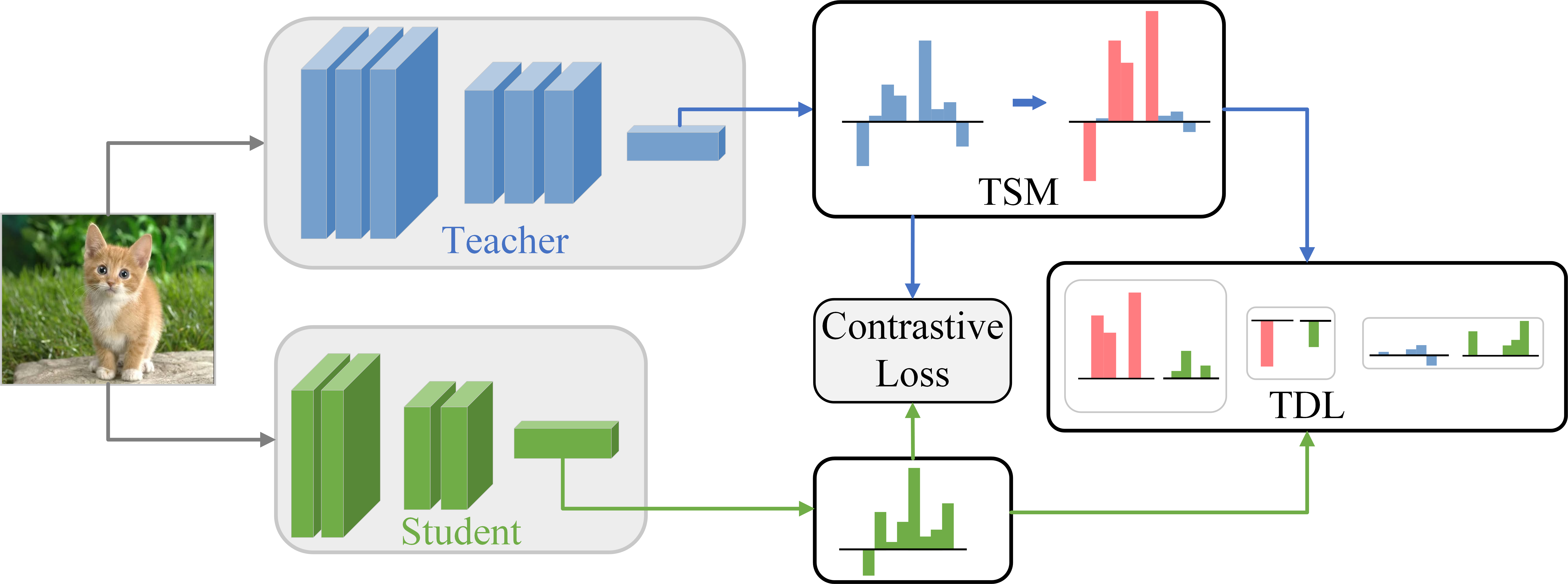}
	\caption{Overall structure of our \textbf{TopKD}.}
	\label{fig:tkd}
\end{figure}

Building upon this insight, we revisit the distribution of teacher's logits and identify the underexplored \textbf{Top-K knowledge}, which captures particularly informative supervision signals. To exploit this property, we propose \textbf{Top-scaled Knowledge Distillation (TopKD)}, a framework that explicitly emphasizes the teacher’s Top-K knowledge during distillation process.\par
Our contributions can be summarized as:
\begin{itemize}
	\item We introduce a novel perspective by analyzing the limitations of conventional logit-based distillation and highlighting the previously overlooked importance of \textbf{Top-K knowledge} within the teacher's logits.
	\item To effectively leverage this insight, we develop \textbf{Top-scaled Knowledge Distillation (TopKD)}, which incorporates a lightweight \textbf{Top-K Scaling Module (TSM)} and a \textbf{Top-K Decoupled Loss (TDL)}. These components are architecture-agnostic and can be seamlessly integrated into existing distillation methods, consistently improving their performance. And, we avoid using KL-Div as the primary loss and instead adopt a contrastive loss to better capture the structure information of the Top-K knowledge.
	\item Experimental results across multiple datasets show that \textbf{TopKD} consistently outperforms both logit-based and feature-based distillation methods, achieving state-of-the-art performance.
\end{itemize}
\section{Related Work}
\label{related_work}
\subsection{Knowledge Distillation}
Knowledge Distillation (KD) transfers knowledge from a large teacher model to a compact student model and has become a widely adopted approach for model compression and performance enhancement. Current KD methods are broadly categorized into logit-based distillation and feature-based distillation. Logit-based distillation \cite{kd, logit3, dkd, dot, ttm} transfers knowledge by aligning the teacher’s and student’s output logits, often using Kullback-Leibler divergence as the objective. Although simple and computationally efficient, these methods often fail to adequately capture the teacher’s internal representational structure. Feature-based distillation \cite{similarity, ab, factor, review, reuse, fcfd} instead guide the student to mimic intermediate feature representations of the teacher, offering richer supervision but at the cost of architectural constraints and increased computational overhead. Additionally, several methods incorporate contrastive learning into the distillation process \cite{crd, wocrd, dcd, gacd}, aiming to enhance the discriminative ability of the student by encouraging it to bring representations closer to the teacher (positive pairs) while distancing them from unrelated samples (negative pairs).\par
\subsection{Contrastive Learning}
Contrastive learning is widely recognized as an effective technique for representation learning, particularly in self-supervised and semi-supervised scenarios. Its core objective is to learn feature embeddings by contrasting positive pairs (e.g., augmented views of the same image) against negative ones (e.g., instances from different classes). Prominent frameworks such as SimCLR \cite{simclr}, MoCo \cite{moco}, and InfoNCE \cite{infonce} demonstrate that enforcing instance-level discrimination leads to robust and transferable feature representations. In the context of knowledge distillation, contrastive losses offer an effective mechanism to encode relational knowledge, complementing traditional instance-level alignment and further enhancing student generalization.\par
\section{Methodology}
\label{methodology}
In this section, we introduce our Top-scaled Knowledge Distillation (TopKD), which aims to enhance logit-based distillation by explicitly leveraging the distribution of the teacher's output. By thoroughly analyzing the distribution of logits, we find that the Top-K logits contain richer and more semantically meaningful supervision signals. However, conventional methods typically rely on KL-Div to force alignment of the entire output distribution, thus neglecting the semantic relevance inherent in the Top-K logits. To address this limitation, we adopt a contrastive loss that captures the relational semantic information of class predictions more effectively. Additionally, we propose two core components: (1) a Top-K Scaling Module (TSM), which adaptively emphasizes the most informative logits during training; and (2) a Top-K Decoupled Loss (TDL), which guides alignment in an embedding space informed by the Top-K logits.\par
\subsection{Review Teacher’s Logits}
\begin{figure}[htbp]
	\centering
	\includegraphics[width=\textwidth]{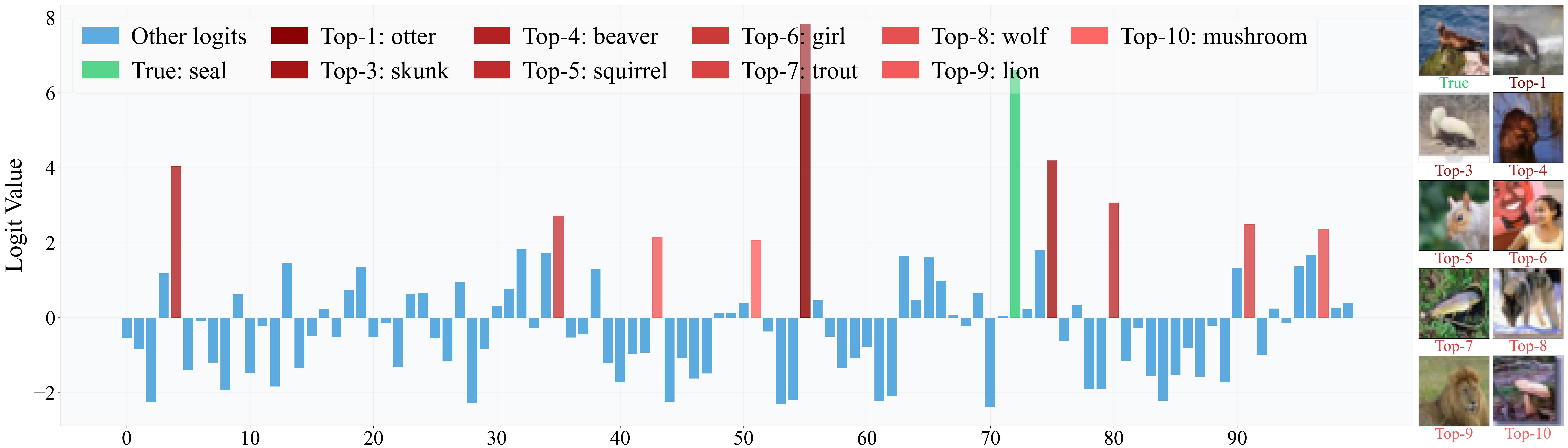}
	\caption{An example from the CIFAR-100 training set where the teacher incorrectly predicts otter for an image labeled seal. While the prediction is incorrect, the Top-10 predictions (highlighted in red) include semantically similar classes, suggesting that the teacher captures meaningful semantic relationships beyond the ground truth.}
	\label{fig:teacher_logits_top10}
\end{figure}
We revisit the teacher model’s logits distribution during training and observe that its top-1 predictions are not always accurate. As shown in Figure~\ref{fig:teacher_logits_top10}, for an input labeled as $\mathit{seal}$, the teacher predicts $\mathit{otter}$, a visually similar but incorrect class. Instead of being purely erroneous, these misclassifications often reflect the semantic hierarchy learned by the teacher. Further inspection of the Top-K predictions (e.g., K=10) reveals that many high-ranking classes are semantically related to the ground truth, indicating that the Top-K logits encode rich relational knowledge. However, as pointed out in~\cite{wasserstein}, conventional distillation based on KL-Div is fundamentally limited: it only encourages alignment of output probabilities between teacher and student, without capturing the underlying inter-class structure. To effectively leverage cross-category information embedded in teacher's logits, we introduce a contrastive learning strategy to align relational patterns between teacher and student representations. Even when the teacher’s top prediction is incorrect, as in Figure~\ref{fig:teacher_logits_top10}, this approach allows the student to extract meaningful semantic relationships, thereby enhancing its generalization ability.\par

To implement this, we adopt a contrastive loss that encourages alignment between semantically corresponding teacher-student pairs while discouraging mismatched pairs. The objective is to pull together the logits of the same samples and push away those of different samples. Let \( \mathbf{z}_s \) and \( \mathbf{z}_t \) denote the student and teacher logits for each batch samples, respectively. The contrastive loss is defined as follows:
\begin{gather}
	\mathcal{L}_{\mathrm{Contrastive}} = \frac{1}{2} \left[
	\mathcal{L}_{\mathrm{CE}}((\mathbf{z}_s * \mathbf{z}_t^\top) / \tau, y) + 
	\mathcal{L}_{\mathrm{CE}}((\mathbf{z}_t * \mathbf{z}_s^\top) / \tau, y)
	\right]
	\label{eq:contrastive_loss}
\end{gather}
where \(\mathcal{L}_{\mathrm{CE}}\) is Cross Entropy (CE) loss, \(\mathbf{z}_s\), \( \mathbf{z}_t\) $\in \mathbb{R}^{B \times C}$, \(B\) represents the batch size and \(C\) is class number, \(\tau\) is a temperature parameter that controls the sharpness of the similarity scores, following the common practice in prior works~\cite{simclr, moco}. Here, $y \in \mathbb{R}^{B \times 1}$, contains indices ranging from 0 to B$-$1. This formulation encourages maximizing similarity for matching pairs along the diagonal of \(\mathbf{z}_s * \mathbf{z}_t^\top\), while minimizing similarity for non-matching pairs off the diagonal.
\subsection{Top-K Scaling Module (TSM)}
To better leverage the cross-category relationships embedded within the teacher’s logits during contrastive learning, we introduce a scaling strategy specifically targeting the Top-K logits. Specifically, we increase the values of the Top-K predictions to emphasize their influence. Additionally, when the teacher's Top-1 prediction is incorrect, we apply a larger scaling factor to the logits of the ground-truth to correct the teacher's output bias. Formally, we rescale the logits for each sample as follows:
\begin{equation}
	z_i' = 
	\begin{cases}
		z_i \cdot w_i + \Delta, & \text{if } i \in \mathcal{I}_\mathrm{top} \cup \{y_g\} \\
		z_i, & \text{otherwise}
	\end{cases}
	\label{eq:logit_scaling}
\end{equation}
where \( \mathcal{I}_\mathrm{top} \) denotes the indices of the Top-K predictions, \( y_g \) is the ground-truth label, \( w_i \) is a rank-dependent scaling factor, and \( \Delta \) is a bias term proportional to the mean difference between Top-K and Non-Top-K logits. The TSM enhances the relative importance of semantically relevant categories while down-weighting noisy or misleading logits, thus more effectively preserving the semantic structure learned by the teacher. Due to space constraints, the detailed procedure for computing \( w_i \) and \( \Delta \) is provided in Algorithm~\ref{algorithm_1} in the appendix.\par
\subsection{Top-K Decoupled Loss (TDL)}
\begin{figure}[htbp]
	\centering
	\begin{minipage}{0.32\textwidth}
		\centering
		\includegraphics[width=\textwidth]{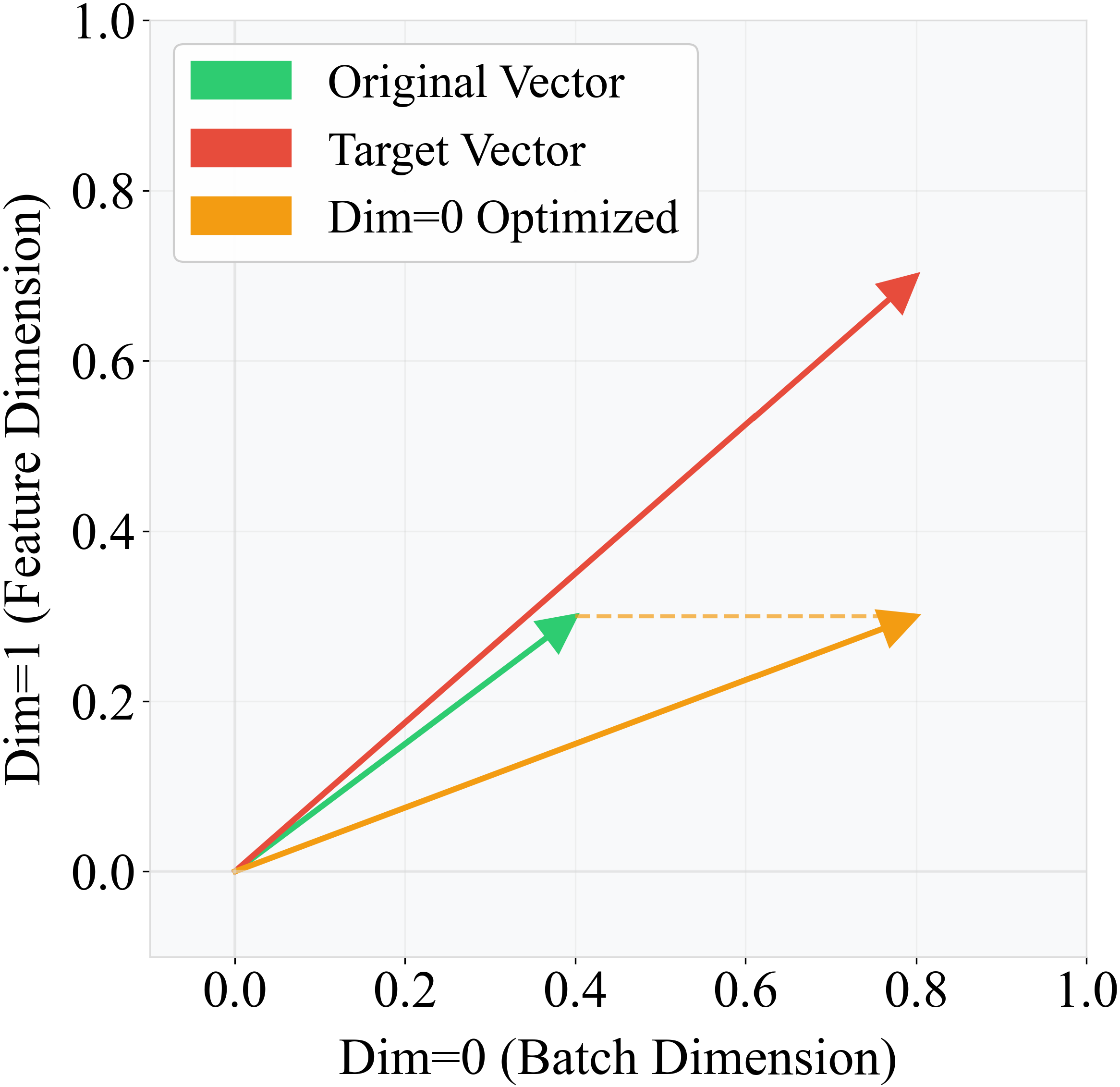}
		\caption*{(a) Dim=0 Only}
	\end{minipage}
	\begin{minipage}{0.32\textwidth}
		\centering
		\includegraphics[width=\textwidth]{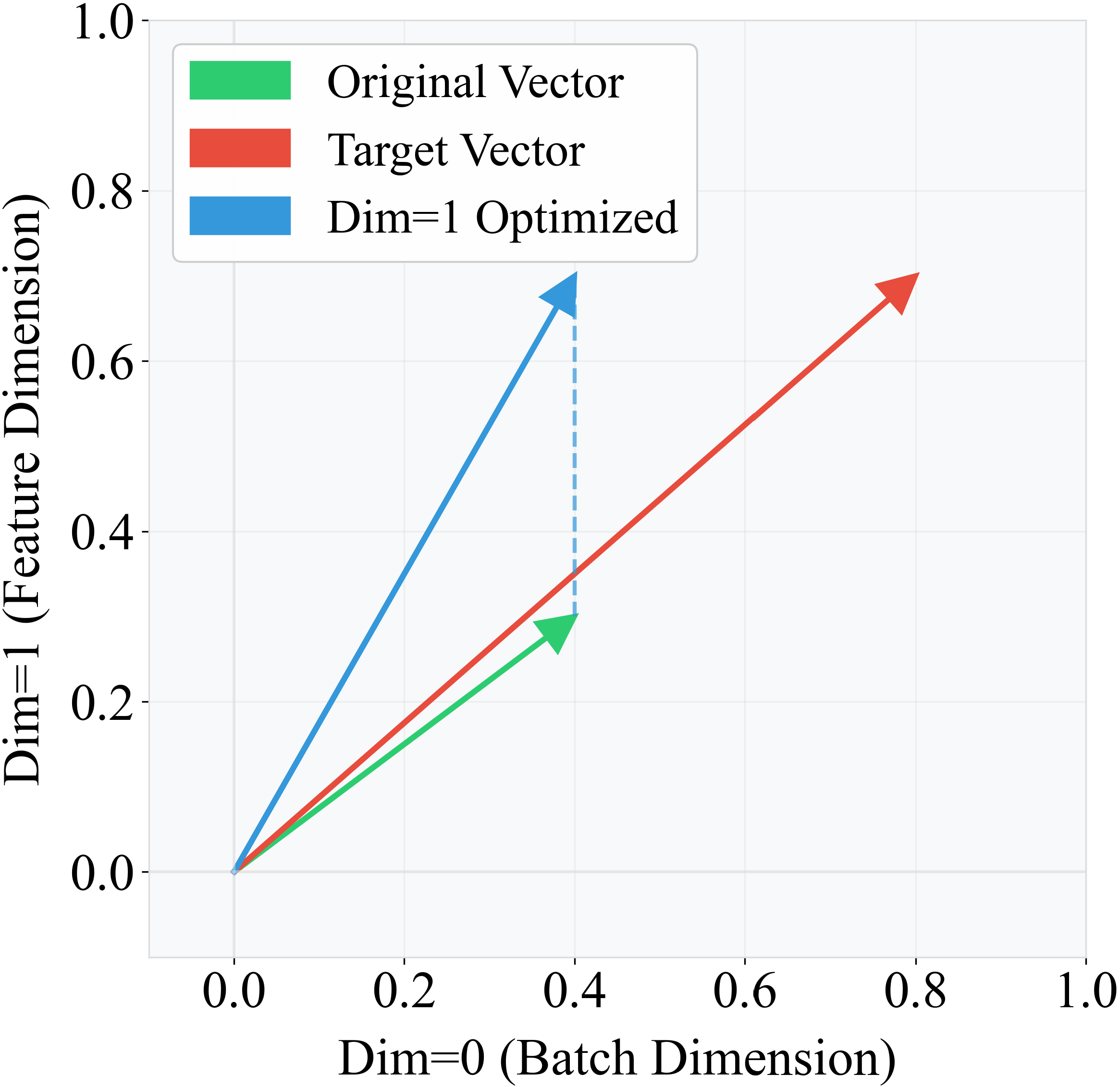}
		\caption*{(b) Dim=1 Only}
	\end{minipage}
	\begin{minipage}{0.32\textwidth}
		\centering
		\includegraphics[width=\textwidth]{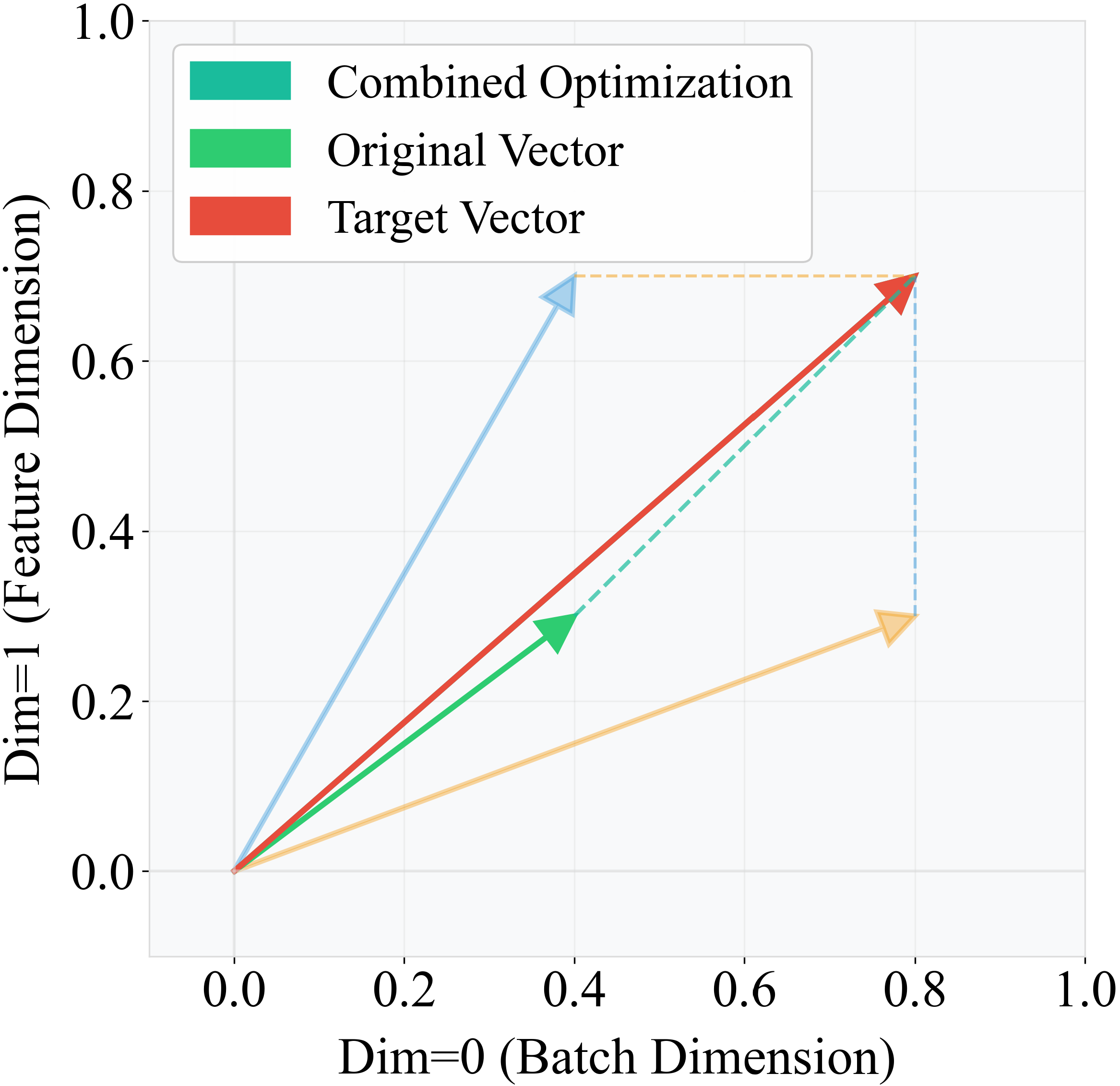}
		\caption*{(c) Dim=0 \& Dim=1 (TopKD)}
	\end{minipage}
	\caption{Comparison of contrastive loss (Dim=0) and cosine similarity loss (Dim=1) for student-teacher alignment. The former aligns instance-wise predictions, while the latter preserves intra structure.}
	\label{fig:logit-decouple}
\end{figure}
The contrastive loss defined in Equation~\ref{eq:contrastive_loss} enforces instance-level alignment by matching each student's logits with its corresponding teacher's logits along the batch dimension (Dim=0 in Figure~\ref{fig:logit-decouple}). However, this objective alone does not capture finer-grained structural discrepancies within individual predictions. To overcome this limitation, we integrate a cosine similarity loss to preserve intra-instance semantic coherence (Dim=1 in Figure~\ref{fig:logit-decouple}). Given two vectors 
\(\mathbf{z}_s\) and \( \mathbf{z}_t\) from the same sample (along feature dimension 
$\mathit{j}$), cosine similarity is computed as:
\begin{gather}
	\text{cos\_sim}(\mathbf{z}_s, \mathbf{z}_t) = \frac{\mathbf{z}_s \cdot \mathbf{z}_t}{\|\mathbf{z}_s\|_2 \cdot \|\mathbf{z}_t\|_2} 
	= \frac{\sum_{j} (\mathbf{z}_{s,i})_{j} \cdot (\mathbf{z}_{t,i})_{j}}{\sqrt{\sum_{j} (\mathbf{z}_{s,i})_{j}^2} \cdot \sqrt{\sum_{j} (\mathbf{z}_{t,i})_{j}^2}}
	\label{eq:normal_cos}
\end{gather}
As shown in Figure~\ref{fig:logit-decouple}(c), Equation~\ref{eq:normal_cos} serves as a natural complement to Equation~\ref{eq:contrastive_loss}, which optimize the optimization direction in the embedding space to better support instance-level alignment. Together, these two objectives ensure semantic consistency and directional guidance, resulting in more discriminative representations.\par

While the composite cosine similarity provides an overall measure of agreement between student and teacher predictions, it does not measure the semantic importance of the different categories in the logits. In practice, a small subset of high-confidence entries, typically the Top-K logits, encodes the most confident and informative information, while the remaining entries may be noisy or ambiguous. Uniformly treating all logits may dilute meaningful semantic signals and hinder effective knowledge transfer, as also observed in \cite{dkd}. Unlike \cite{dkd}, which employs a KL-Div variant, our method leverages cosine similarity. These two approaches reflect fundamentally different alignment principles: KL-Div enforces exact value matching, while cosine similarity emphasizes directional consistency in the embedding space.\par

To better exploit logits, we decouple global cosine similarity into three distinct components based on the sign and magnitude of teacher logits: (1) Positive Top-K, corresponding to the dimensions with the highest positive values; (2) Negative Top-K, covering the most negative values; and (3) Non-Top-K, representing the remaining, less confident dimensions. The Top-K Decoupled Loss as follows:
\begin{equation}
	\mathcal{L}_{\mathrm{TDL}} = 1 - \frac{1}{B} \sum_{i=1}^{B} \Bigg[
	\alpha * \cos\Big( \mathbf{z}_{s,i}^{\text{Pos}}, \mathbf{z'}_{t,i}^{\text{Pos}} \Big) +
	\beta *  \cos\Big( \mathbf{z}_{s,i}^{\text{Neg}}, \mathbf{z'}_{t,i}^{\text{Neg}} \Big) +
	\cos\Big( \mathbf{z}_{s,i}^{\text{Non}}, \mathbf{z'}_{t,i}^{\text{Non}} \Big)
	\Bigg]
\end{equation}
where \(B\) is the batch size, \(cos\) denotes the cosine similarity defined in Equation~\ref{eq:normal_cos}, and \(\mathbf{z'}_{t}\) represents teacher's logits after applying the scaling in Equation~\ref{eq:logit_scaling}. Only the teacher’s logits are scaled; the student’s logits remain unchanged. We perform a detailed ablation of \(\alpha\) and \(\beta\) in Table~\ref{tab:ablation-alpha}.
\paragraph{TopKD Loss.} The final TopKD loss function is defined as:
\begin{equation}
	\mathcal{L}_{\text{TopKD}} = \mathcal{L}_{\mathrm{Contrastive}} + \mathcal{L}_{\mathrm{TDL}}
\end{equation}
In TopKD, the overall distillation objective consists of two complementary components: a contrastive loss that enforces instance-level alignment between student and teacher predictions, and the Top-K Decoupled Loss (TDL), which captures fine-grained structural consistency in logits. 
\section{Experiments}
\label{experiments}
\paragraph{Dataset.} Our experiments primarily focus on image classification. CIFAR-100 \cite{cifar100} is a widely used benchmark with 60,000 color images (32×32), split into 50,000 training and 10,000 validation images across 100 classes. ImageNet \cite{imagenet} is a large-scale dataset comprising 1.28 million training and 50,000 validation images over 1,000 categories. STL-10 \cite{stl-10} and Tiny-ImageNet \cite{tiny-imagenet} serve as mid-scale benchmarks for evaluating generalization. STL-10 provides 10 high-resolution classes (96×96) and a large unlabeled set for unsupervised learning, while Tiny-ImageNet includes 200 low-resolution classes (64×64), designed to test generalization under limited data per class.\par
\paragraph{Baselines.} In recent years, most state-of-the-art methods have relied on feature-based distillation. Thus, we first compare TopKD with representative feature-based methods, including CRD \cite{crd}, ReviewKD \cite{review}, SimKD \cite{reuse}, CAT-KD \cite{class} and FCFD \cite{fcfd}. Since TopKD is a logit-based method, we also compare it with classic KD \cite{kd} and recent logit-based distillation methods, including CTKD \cite{ctkd}, DKD \cite{dkd}, DOT \cite{dot}, LSKD \cite{lskd}, and WTTM \cite{ttm}. We report LSKD \cite{lskd}, select DKD+LS \cite{dkd,lskd} as the representative setting, since MLKD+LS \cite{mlkd,lskd} uses more epochs and different initial learning rates. We selected ResNets \cite{resnet}, WRNs \cite{wrn}, VGGs \cite{vgg}, MobileNets \cite{mobilenet}, and ShuffleNets \cite{shufflenet} as the baseline models for our classification experiments on CIFAR-100 \cite{cifar100} and ImageNet \cite{imagenet} datasets. For Vision Transformer baselines, including DeiT-Ti \cite{deti}, T2T-ViT7 \cite{t2t}, PiT-Ti \cite{pit}, and PVT-Ti \cite{pvt}.\par
\paragraph{Training details.} We adopt the experimental settings from prior work \cite{crd,dkd,review}. For all experiments, we used the SGD \cite{sgd} optimizer with a batch size of 64, training for 240 epochs on CIFAR-100 and 100 epochs on ImageNet. For STL-10 \cite{stl-10} and Tiny-ImageNet \cite{tiny-imagenet}, we performed transfer learning by freezing the intermediate layers and training only the classifier to evaluate representation transferability. More implementation details are provided in the appendix.\par
\subsection{Main Results}
\label{main_results}
\paragraph{CIFAR-100 Image Classification.} We evaluate various KD methods on CIFAR-100, using heterogeneous teacher-student architectures (Table~\ref{tab:cifar100-dif-table}) and homogeneous ones (Table~\ref{tab:cifar100-same-table}). Methods are categorized by distillation type: feature-based and logit-based. TopKD consistently perform best or second-best, demonstrating its effectiveness. Furthermore, Using the same settings as MLKD+LS \cite{mlkd,lskd}, our method achieves higher accuracy. Detailed results are in the appendix (Table~\ref{tab:tkd-hyperparam-combined}).\par
\begin{table}[htbp]
	\centering
	\caption{The Top-1 accuracy (\%) of knowledge distillation methods on the CIFAR-100 \cite{cifar100} validation set. The teacher and student models have a \textbf{heterogeneous structure}. Methods marked with * indicate reproduced results under the same hardware and experimental conditions. All results are reported as the average over three trials.}
	\label{tab:cifar100-dif-table}
	\resizebox{\textwidth}{!}{%
		\begin{tabular}{@{}clccccccc@{}}
			\toprule
			& \multirow{2}{*}{Teacher} & ResNet32×4    & ResNet32×4 & ResNet32×4 & WRN-40-2  & WRN-40-2     & VGG13        & ResNet50     \\
			Distillation              &            & 79.42 & 79.42 & 79.42 & 75.61 & 75.61 & 74.64 & 79.34 \\
			Manner & \multirow{2}{*}{Student} & SHN-V2 & WRN-16-2   & WRN-40-2   & ResNet8×4 & MN-V2 & MN-V2 & MN-V2 \\
			&            & 71.82 & 73.26 & 75.61 & 72.50 & 64.60 & 64.60 & 64.60 \\ \midrule
			& CRD \cite{crd}       & 75.65 & 75.65 & 78.15 & 75.24 & 70.28 & 69.73 & 69.11 \\
			& ReviewKD \cite{review}   & 77.78 & 76.11 & 78.96 & 74.34 & \underline{71.28} & 70.37 & 69.89 \\
			\multirow{2}{*}{Feature}
			& SimKD \cite{reuse}     & 78.39 & \underline{77.17} & \underline{79.29} & 75.29 & 70.10 & 69.44 & 69.97 \\
			& CAT-KD \cite{class}    & \underline{78.41} & 76.97 & 78.59 & 75.38 & 70.24 & 69.13 & \underline{71.36} \\
			& FCFD* \cite{fcfd}  & 78.25 & 76.73 & 79.27 & 76.26 & n/a & \underline{70.69} & 70.89 \\
			\midrule
			& KD \cite{kd}        & 74.45 & 74.90 & 77.70 & 73.07 & 68.36 & 67.37 & 67.35 \\
			& DKD \cite{dkd}       & 77.07 & 75.70 & 78.46 & 75.56 & 69.28 & 69.71 & 70.35 \\
			\multirow{2}{*}{Logit}   
			& DKD+LS* \cite{lskd} & 77.09 & 75.75 & 78.60 & 75.93  & 69.28 & 69.98 & 70.78 \\
			\multicolumn{1}{l}{}      
			& WTTM* \cite{ttm}    & 76.59 & 76.04 & 78.45 & \underline{76.40} & 68.81 & 69.18 & 69.95 \\
			& \textbf{TopKD}       & \textbf{78.83} & \textbf{77.47} & \textbf{80.09} & \textbf{76.88} & \textbf{71.42} & \textbf{71.60} & \textbf{72.54} \\ 
			\bottomrule
		\end{tabular}%
	}
\end{table}
\begin{table}[htbp]
	\centering
	\caption{Top-1 accuracy (\%) of knowledge distillation methods on the CIFAR-100 \cite{cifar100} validation set. The teacher and student models have a \textbf{homogeneous structure}. All results are reported as the average over three trials.}
	\label{tab:cifar100-same-table}
	\resizebox{\textwidth}{!}{%
		\begin{tabular}{@{}clccccccc@{}}
			\toprule
			& \multirow{2}{*}{Teacher} & ResNet32×4 & VGG13 & WRN-40-2 & WRN-40-2 & ResNet56 & ResNet110 & ResNet110 \\
			Distillation              &            & 79.42 & 74.64 & 75.61 & 75.61 & 72.34 & 74.31 & 74.31 \\
			Manner & \multirow{2}{*}{Student} & ResNet8x4  & VGG8  & WRN-40-1 & WRN-16-2 & ResNet20 & ResNet32  & ResNet20  \\
			&            & 72.50 & 70.36 & 71.98 & 73.26 & 69.06 & 71.14 & 69.06 \\ \midrule
			& CRD \cite{crd}       & 75.51 & 73.94 & 74.14 & 75.48 & 71.16 & 73.48 & 71.46 \\
			& ReviewKD \cite{review}  & 75.63 & 74.84 & \underline{75.09} & 76.12 & 71.89 & 73.89 & 71.34 \\
			\multirow{2}{*}{Feature}
			& SimKD \cite{reuse}     & \underline{78.08} & 74.80 & 74.53 & 75.53 & 71.05 & 73.92 & 71.06 \\
			& CAT-KD \cite{class}    & 76.91 & 74.65 & 74.82 & 75.60 & 71.62 & 73.62 & 71.37 \\  
			& FCFD* \cite{fcfd}        & 76.98 & \underline{74.96} & \textbf{75.53} & \textbf{76.34} & 71.68 & 73.75 & \underline{71.92} \\
			\midrule
			& KD \cite{kd}        & 73.33 & 72.98 & 73.54 & 74.92 & 70.66 & 73.08 & 70.67 \\
			& DKD \cite{dkd}       & 76.32 & 74.68 & 74.81 & 76.24 & 71.97 & 74.11 & 71.06 \\
			\multirow{2}{*}{Logits}   
			& DKD+LS* \cite{lskd}  & 76.79 & 74.80 & 74.73 & \textbf{76.34} & \textbf{72.31} & 73.89 & 71.39 \\
			\multicolumn{1}{l}{} 
			& WTTM* \cite{ttm}    & 76.24 & 74.27 & 74.08 & 76.29 & 71.65 & \underline{74.26} & 71.16 \\
			& \textbf{TopKD}       & \textbf{78.18} & \textbf{75.19} & \underline{75.09} & \underline{76.32} & \underline{71.99} & \textbf{74.79} & \textbf{72.19} \\
			\bottomrule
		\end{tabular}%
	}
\end{table}
\paragraph{ImageNet Image Classification.} We report Top-1 and Top-5 accuracy in Table~\ref{tab:combined_imagenet}. In the homogeneous setting (ResNet-34 / ResNet-18), the teacher and student models show minimal performance difference. In the heterogeneous setting (ResNet-50 / MobileNetV1), the teacher significantly outperforms the student. Across both settings, TopKD consistently outperforms most state-of-the-art distillation methods on the large-scale ImageNet dataset.\par
\begin{table}[htbp]
	\centering
	\caption{Top-1 and Top-5 accuracy (\%) on the ImageNet \cite{imagenet} validation set. Distillation is performed under two settings: (a) ResNet-34/ResNet-18 (\textbf{homogeneous}) and (b) ResNet-50/MobileNet-V1 (\textbf{heterogeneous}). Tch. = Teacher; Stu. = Student.}
	\label{tab:combined_imagenet}
	\resizebox{\textwidth}{!}{%
		\begin{tabular}{@{}c|c|cc|ccc|ccccc@{}}
			\toprule
			\multirow[c]{2}{*}{Setting}
			& \multirow[c]{2}{*}{Metric}
			& \multirow[c]{2}{*}{Tch.}
			& \multirow[c]{2}{*}{Stu.}
			& \multicolumn{3}{c|}{Feature}
			& \multicolumn{5}{c}{Logits} \\
			\cmidrule(lr){5-7}\cmidrule(lr){8-12}
			&  &  & 
			& CRD~\cite{crd} & ReviewKD~\cite{review} & FCFD~\cite{fcfd}
			& DKD~\cite{dkd} & DKD+LS~\cite{lskd} & WTTM~\cite{ttm}
			& WKD-L~\cite{wasserstein} & \textbf{TopKD} \\
			\midrule
			\multirow{2}{*}{(a)}
			& Top-1
			& 73.31 & 69.75
			& 71.17 & 71.61 & \underline{72.24}
			& 71.70 & 71.88 & 72.19 & \textbf{72.49} & 71.52 \\
			& Top-5
			& 91.42 & 89.07
			& 90.13 & 90.51 & \underline{90.74}
			& 90.41 & 90.58 & n/a   & \textbf{90.75} & \textbf{90.75} \\
			\midrule
			\multirow{2}{*}{(b)}
			& Top-1
			& 76.16 & 68.87
			& 71.37 & 72.56 & \underline{73.37}
			& 72.05 & 72.85 & 73.09 & 73.17 & \textbf{73.51} \\
			& Top-5
			& 92.86 & 88.76
			& 90.41 & 91.00 & \underline{91.35}
			& 91.05 & 91.23 & n/a   & 91.32 & \textbf{91.95} \\
			\bottomrule
		\end{tabular}%
	}
\end{table}
\subsection{Ablation Studies}
\label{ablation_studies}
To assess the contributions of TopKD's core components, we conduct ablation studies on the Top-K Scaling Module (TSM) and the Top-K Decoupled Loss (TDL). As shown in Table~\ref{tab:ablation-tsm-tdl}, both modules individually improve the baseline, and their combination yields additional gains, suggesting that they offer complementary benefits to overall performance. We further investigate the sensitivity of TopKD to the choice of K in the Top-K selection. Specifically, we evaluate the impact of varying the value of K on overall performance in Table~\ref{tab:ablation-topk}. In this experiment, we use ResNet32×4 \cite{resnet} as the teacher and ResNet8×4 \cite{resnet} as the student.\par
\begin{table}[htbp]
	\centering
	\caption{Ablation study of TopKD on CIFAR-100 \cite{cifar100} validation set. We compare the original KL-Div (Baseline), contrastive loss without TSM and TDL, and selectively enabling TSM and TDL.}
	\label{tab:ablation-tsm-tdl}
	\resizebox{\textwidth}{!}{%
		\begin{tabular}{@{}l|c|c|c|c|>{}c@{}}
			\toprule
			Methods & KL-Div & Contrastive Loss & +TSM only & +TDL only & \textbf{TopKD (TSM + TDL)} \\
			\midrule
			Top-1 Acc (\%)       & 73.33 & 77.65 & 77.83 & 78.09 & \textbf{78.18} \\
			Gain over KL-Div (\%)& --    & +4.32 & +4.50 & +4.76 & \textbf{+5.15} \\
			\bottomrule
		\end{tabular}%
	}
\end{table}
\begin{table}[htbp]
	\centering
	\caption{Ablation study on the effect of different Top-K values. Top-1 accuracy (\%) is reported on the CIFAR-100 \cite{cifar100} validation set. We adopt K=10 for all experiments.}
	\label{tab:ablation-topk}
	\resizebox{\textwidth}{!}{%
	\begin{tabular}{l|cccc|ccc|ccc}
		\toprule
		K Values & 1     & 3     & 5     & \textbf{10}     & 15     & 20     & 25   & 30   & 40   & 50 \\
		\midrule
		Top-1 Acc (\%) & 75.41 & \underline{78.10} & \underline{78.10} & \textbf{78.18} & 78.06 & 77.90 & 77.94 & 77.97 &  77.87 & 60.70 \\
		\bottomrule
	\end{tabular}
	}
\end{table}
The results indicate that more informative knowledge is captured when K=3, 5, 10. Increasing the value of K indefinitely does not lead to more knowledge, but instead only makes Top knowledge less prominent compared to other knowledge. In Table~\ref{tab:ablation-alpha}, We observe that \(\alpha\)=3, 5 yield the best results in both settings, respectively, with similar performance gaps across \(\beta\) values, confirming that Top-K knowledge is most critical. Accordingly, we adopt \(\alpha\)=3 and \(\beta\)=1 as our defaults for all experiments.\par
\begin{table}[htbp]
	\centering
	\caption{Ablation study on the effect of \(\alpha\) and \(\beta\) values in Top-K Decoupled Loss (TDL). We set alpha=5 as the upper limit for positive weights and beta=2 for negative weights, based on the values of contrastive loss and TDL.}
	\label{tab:ablation-alpha}
	\resizebox{\textwidth}{!}{%
		\begin{tabular}{@{}ll|ccccc|ccccc@{}}
			\toprule
			\multicolumn{2}{l|}{\(\beta\) Values}  & \multicolumn{5}{c|}{\(\beta\)=1}      & \multicolumn{5}{c}{\(\beta\)=2}       \\ \midrule
			\multicolumn{2}{l|}{\(\alpha\) Values} & 5.0   & 4.0   & \textbf{3.0}   & 2.0   & 1.0   & \textbf{5.0}  & 4.0   & 3.0   & 2.0   & 1.0   \\ \midrule
			\multicolumn{2}{l|}{Top-1 Acc (\%)}    & 78.19 & 78.00 & \textbf{78.23} & 77.77 & 77.66 & \textbf{78.27} & 78.12 & 77.88 & 77.67 & 77.48 \\ \bottomrule
		\end{tabular}%
	}
\end{table}

\subsection{Extensions}
\label{extensions}
\begin{figure}[htbp]
	\centering
	\begin{minipage}{0.495\textwidth}
		\centering
		\includegraphics[width=\textwidth]{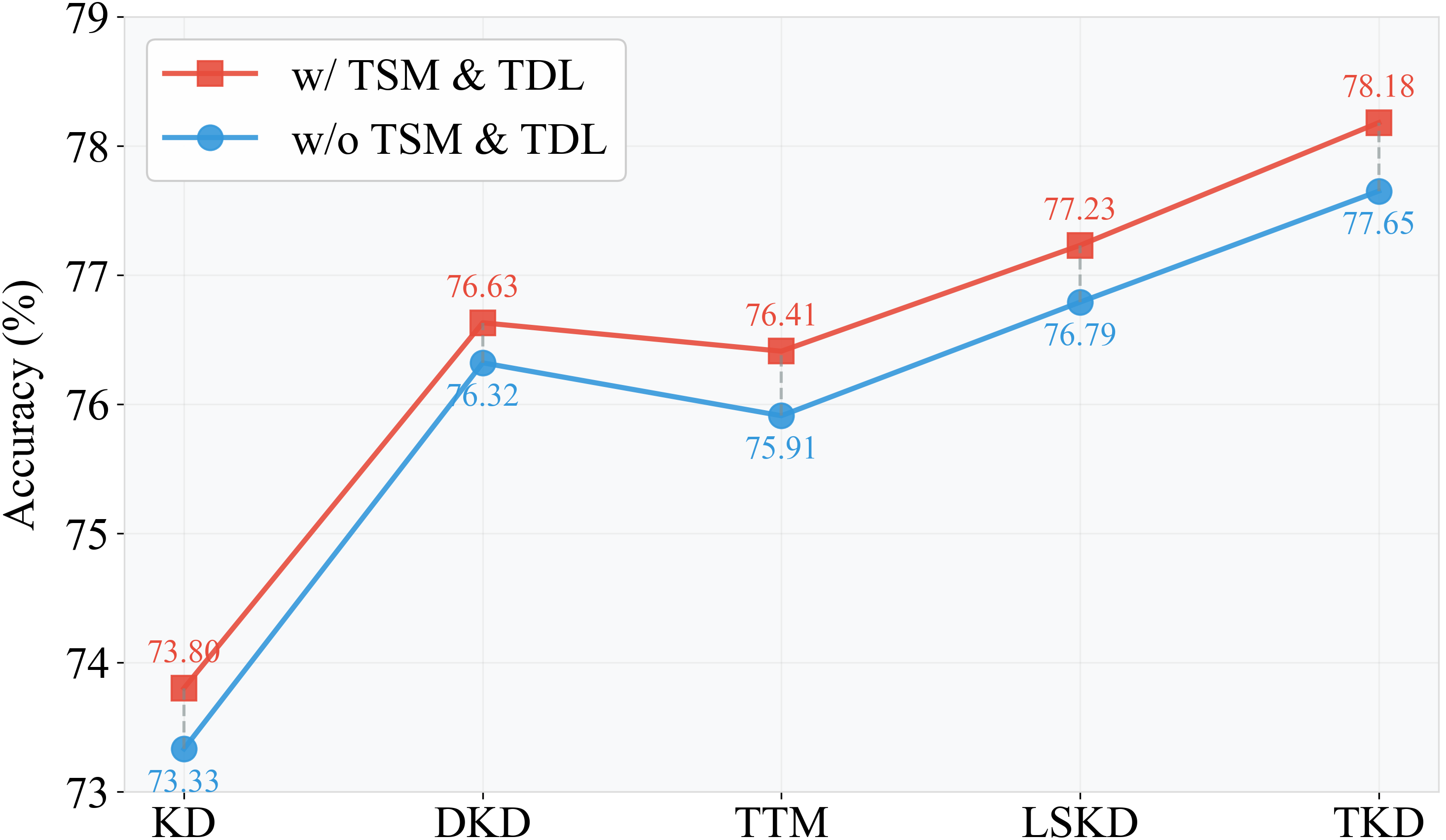}
		\caption*{(a) Logit-Based Distillation Methods}
	\end{minipage}
	\begin{minipage}{0.495\textwidth}
		\centering
		\includegraphics[width=\textwidth]{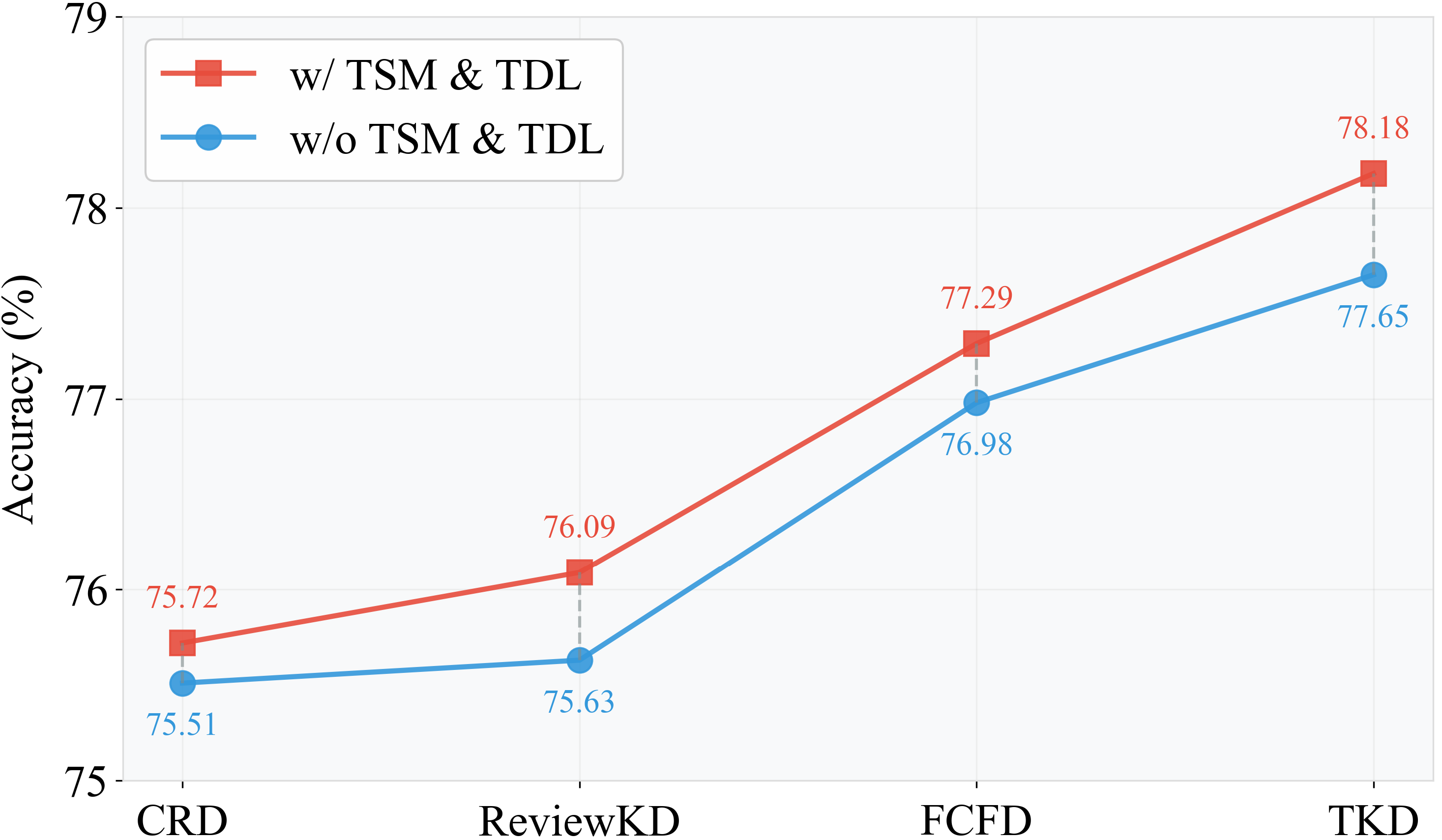}
		\caption*{(b) Feature-Based Distillation Methods}
	\end{minipage}
	\caption{Performance comparison before and after integrating the TSM \& the TDL components into existing distillation methods. Use ResNet32x4 \cite{resnet} as teacher and ResNet8x4 \cite{resnet} as student.}
	\label{fig:logit-feature-comparison}
\end{figure}

To evaluate TopKD, we analyze it from five key perspectives. First, we integrate the plug-and-play TSM and TDL modules into various distillation frameworks, consistently improving performance, demonstrating TopKD's generalizability and compatibility. Second, we test the transferability of the distilled models on downstream datasets, assessing TopKD’s ability to generalize beyond the original domain. Third, we provide visualizations that clearly validate the superiority of TopKD. Fourth, we explore the issue of larger models not always making better teachers and show that TopKD mitigates this, even with a significant teacher-student capacity gap. Finally, we extend the evaluation to Vision Transformer models, confirming TopKD's performance improvements.
\paragraph{Plug-and-Play Capability.} One of the key advantages of TopKD lies in its modular design, which enables seamless integration into existing knowledge distillation methods. Both the Top-K Scaling Module (TSM) and the Top-K Decoupled Loss (TDL) are lightweight and architecture-agnostic, requiring no modifications to the backbone models. To evaluate this capability, we incorporate TSM and TDL into several representative distillation methods. As shown in Figure~\ref{fig:logit-feature-comparison}, each integration leads to consistent and notable performance improvements over the original methods. These results suggest that TSM and TDL serve as general enhancement module, exhibiting strong compatibility, transferability, and robustness across distillation methods.\par
\paragraph{Transferability of representations.} Knowledge distillation can promote the transfer of useful representations to downstream tasks or datasets that differ from the original training domain. To evaluate the transferability of features learned through TopKD, we adopt ResNet8x4 and MobileNetV2 as student models, distilled from ResNet32x4 and ResNet50 respectively, or trained from scratch on CIFAR-100 \cite{cifar100} for comparison. For the STL-10 \cite{stl-10} and Tiny-ImageNet \cite{tiny-imagenet} benchmarks (resized to 32×32), we freeze the model encoder up to the penultimate layer and train a linear classifier on top to perform 10-class and 200-class classification, respectively. As shown in Table~\ref{tab:transfer-merged}, TopKD consistently improves the quality of transferred representations, demonstrating strong generalization across datasets and tasks.\par
\begin{figure}[htbp]
	\centering
	\begin{minipage}{0.19\textwidth}
		\centering
		\includegraphics[width=\textwidth]{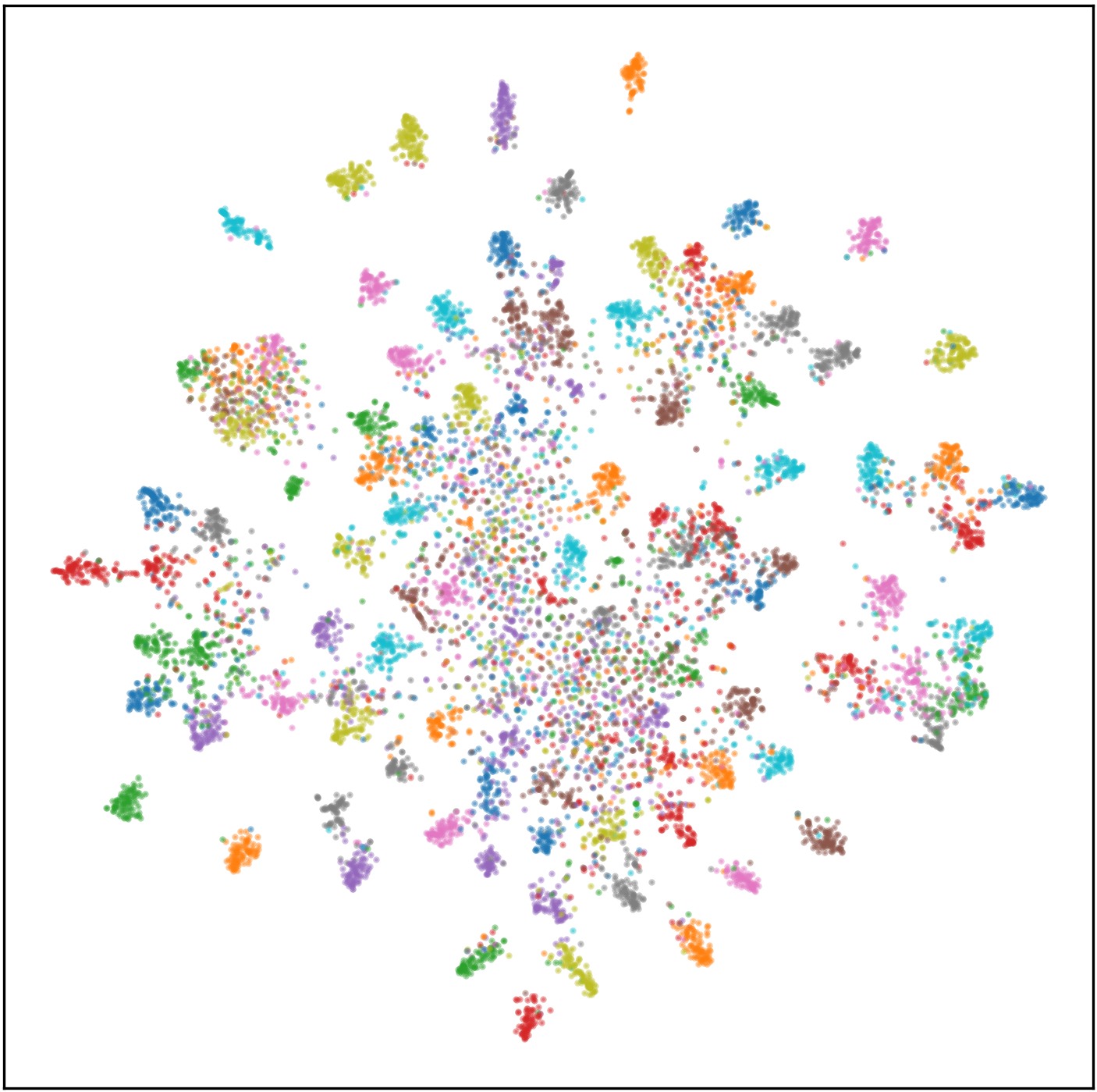}
		\caption*{(a) DOT \cite{dot}}
	\end{minipage}
	\begin{minipage}{0.19\textwidth}
		\centering
		\includegraphics[width=\textwidth]{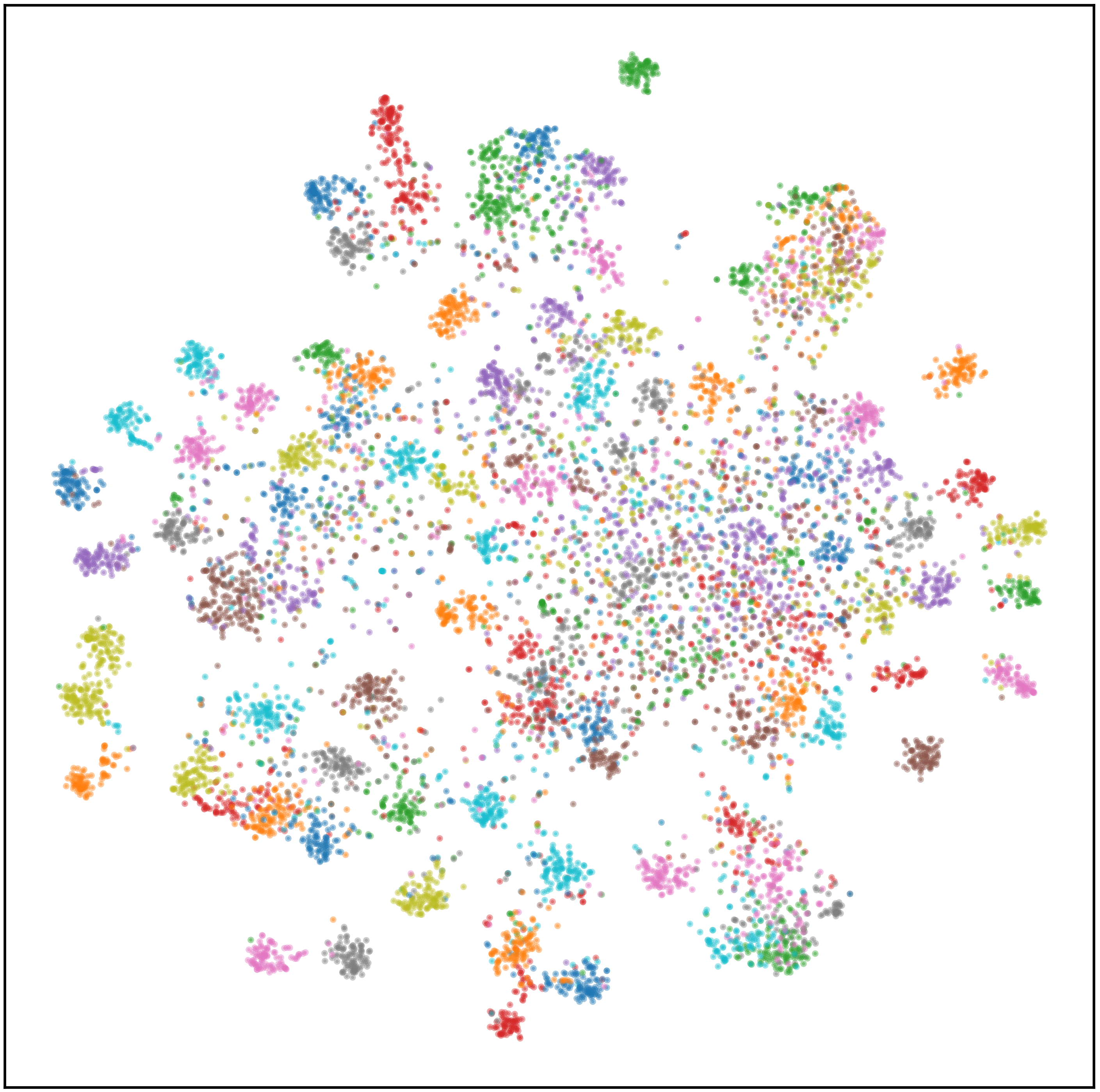}
		\caption*{(b) FCFD \cite{fcfd}}
	\end{minipage}
	\begin{minipage}{0.19\textwidth}
		\centering
		\includegraphics[width=\textwidth]{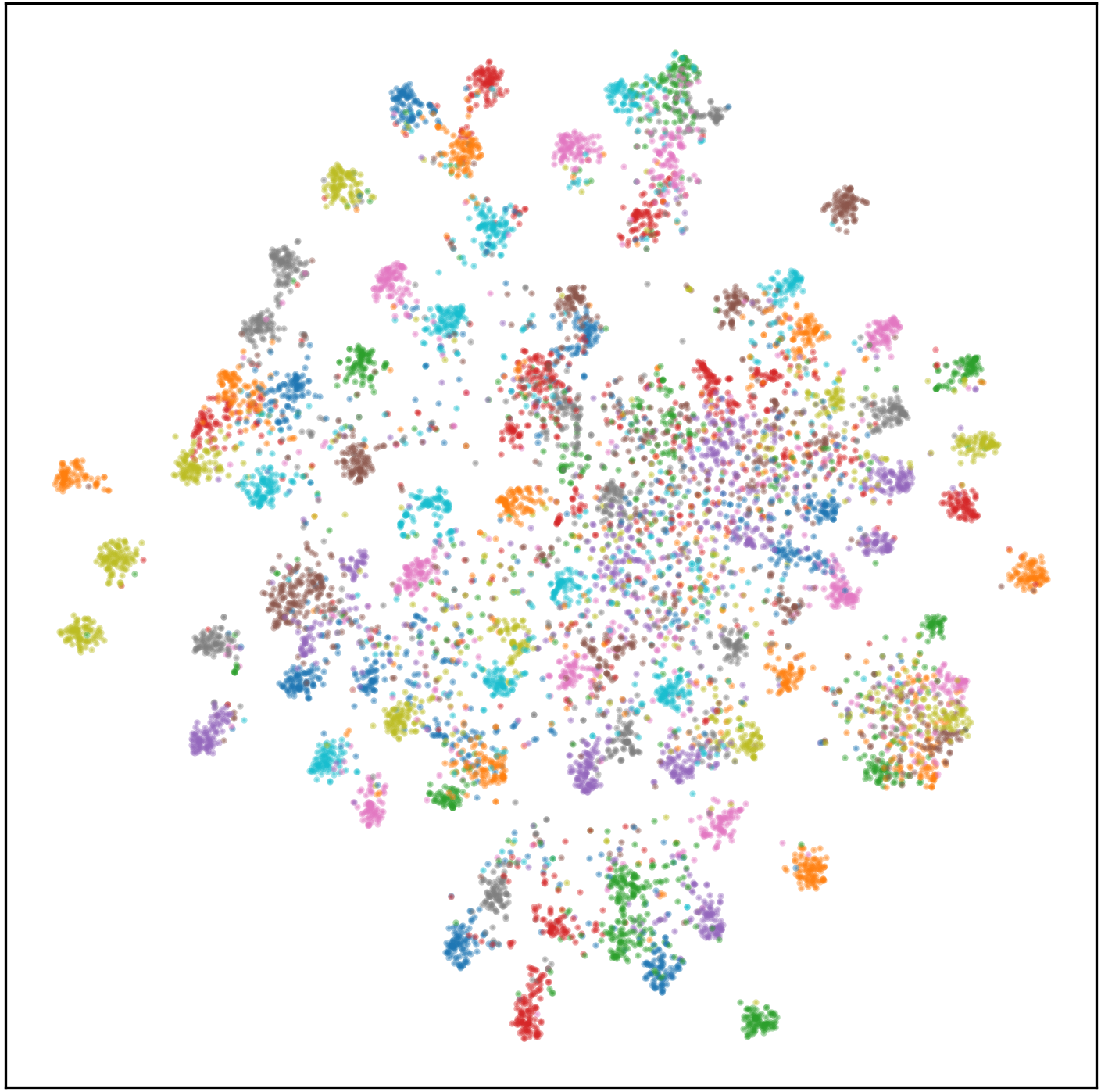}
		\caption*{(c) WTTM \cite{ttm}}
	\end{minipage}
	\begin{minipage}{0.19\textwidth}
		\centering
		\includegraphics[width=\textwidth]{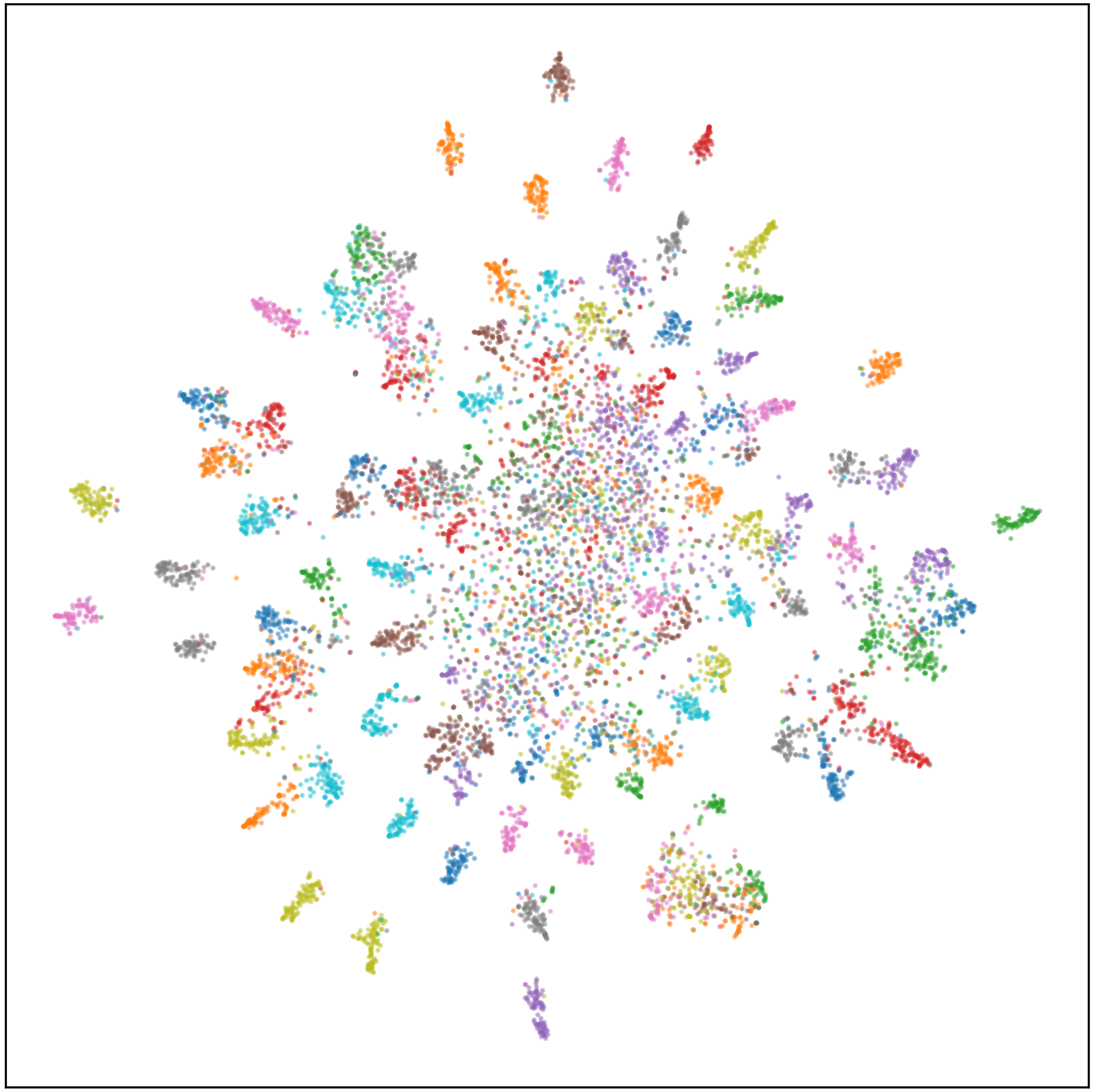}
		\caption*{(d) \textbf{TopKD}}
	\end{minipage}
	\begin{minipage}{0.19\textwidth}
		\centering
		\includegraphics[width=\textwidth]{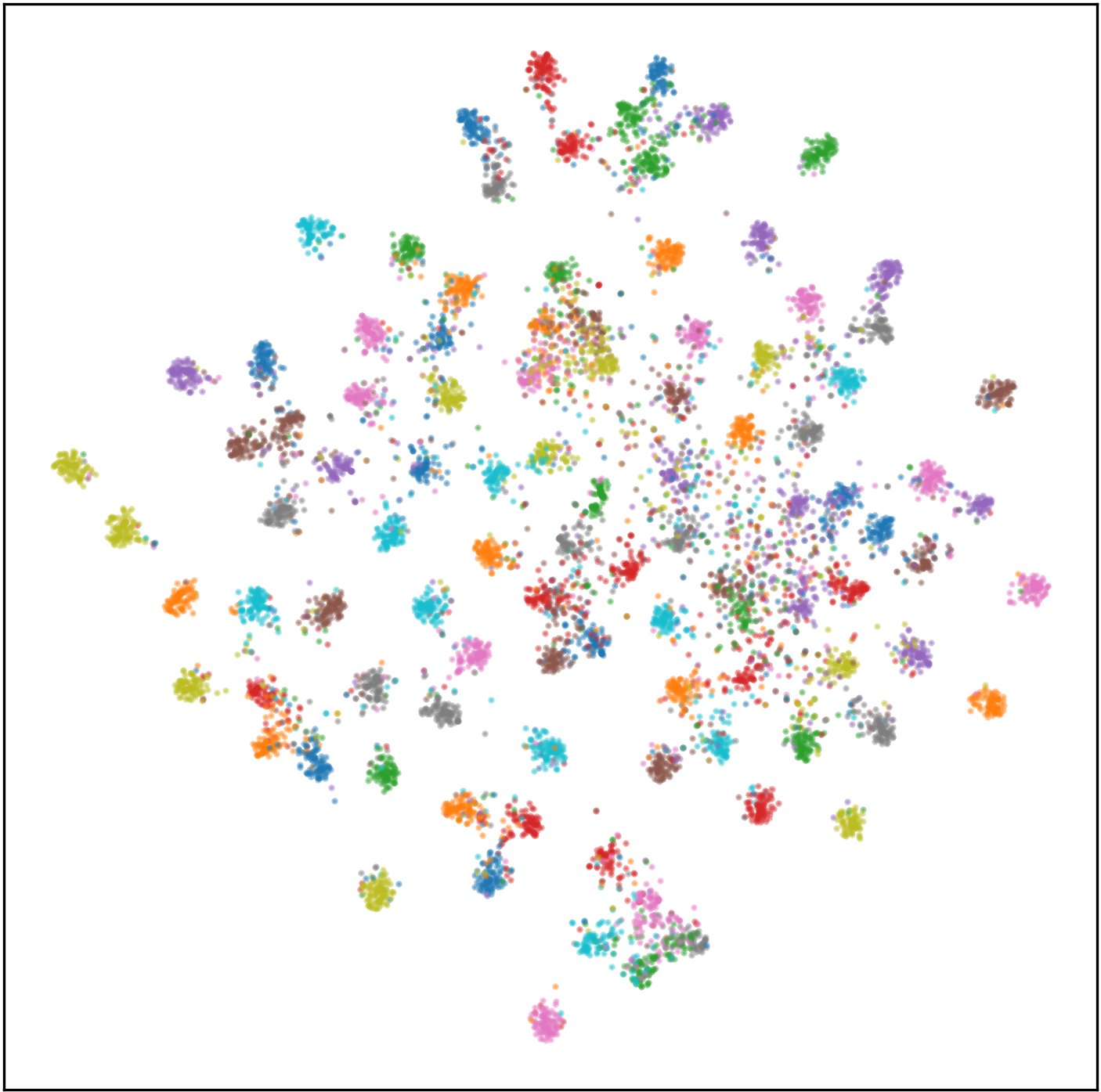}
		\caption*{(e) Teacher}
	\end{minipage}
	\caption{The t-SNE \cite{tsne} visualization of features in a \textbf{homogeneous structure}. Our method yields more separable class distributions.}
	\label{fig:tsne_comparison}
\end{figure}
\begin{figure}[htbp]
	\centering
	\begin{minipage}{0.185\textwidth}
		\includegraphics[width=\textwidth]{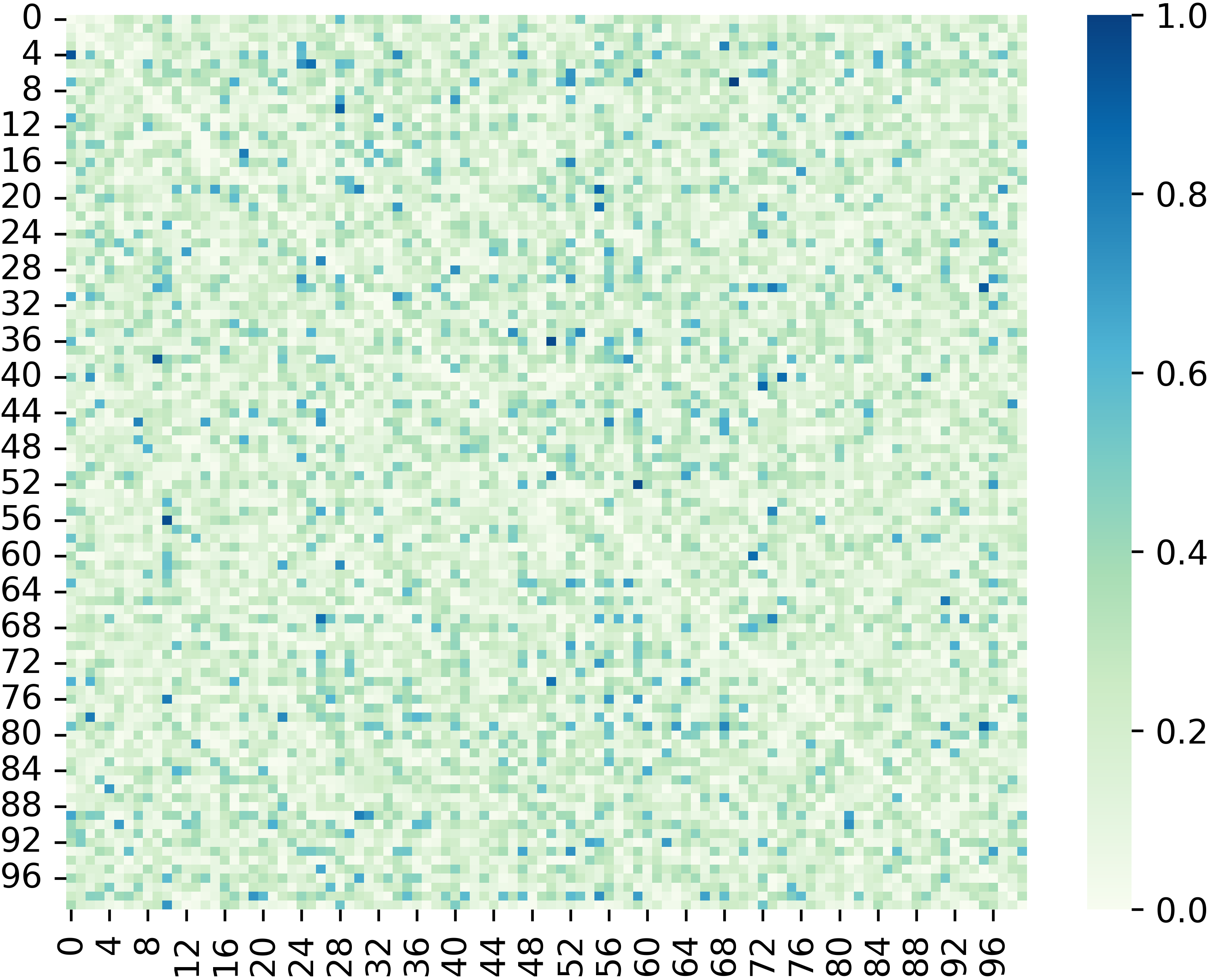}
		\caption*{(a) FCFD \cite{fcfd}}
	\end{minipage}
	\begin{minipage}{0.185\textwidth}
		\includegraphics[width=\textwidth]{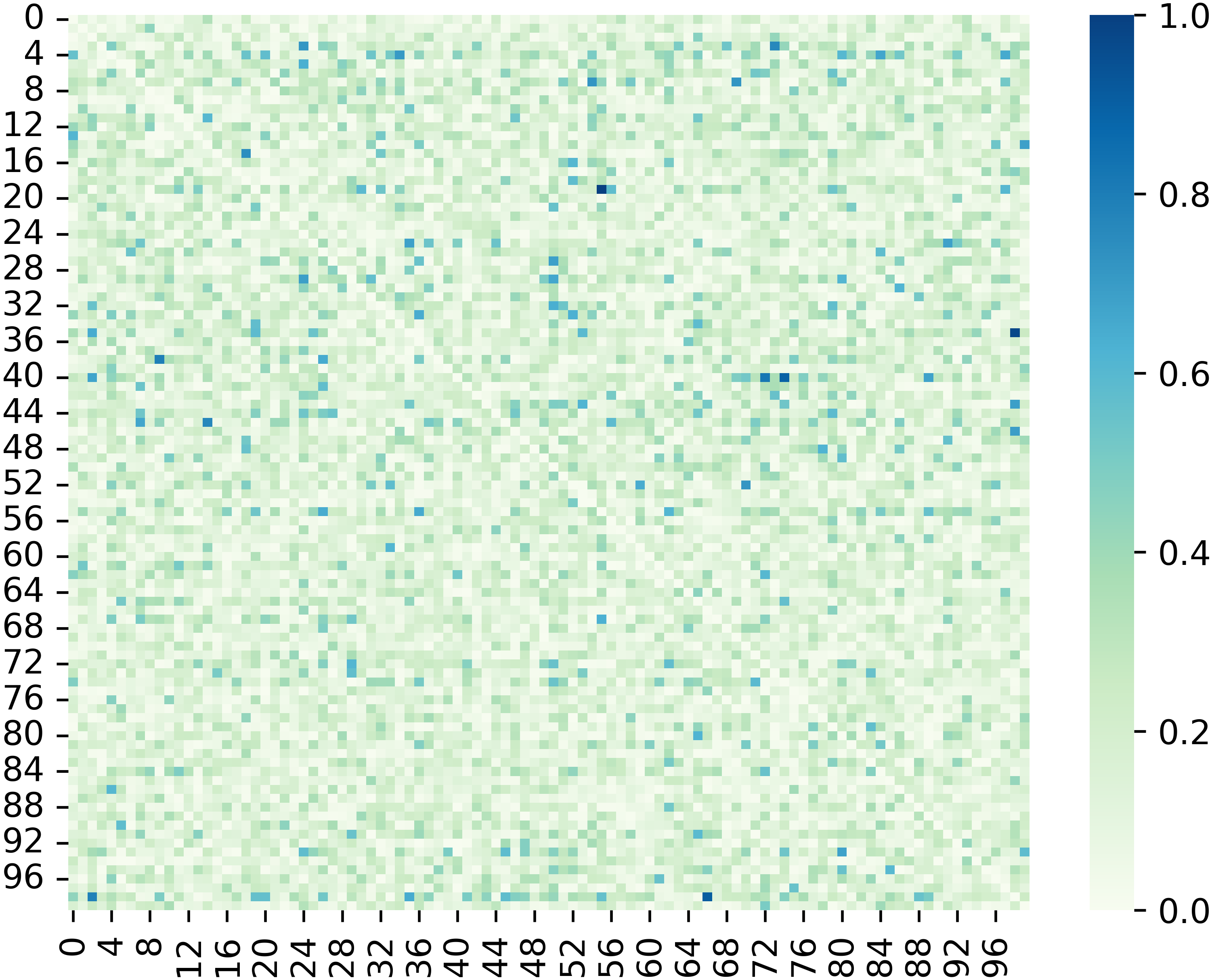}
		\caption*{(b) DKD+LS \cite{lskd}}
	\end{minipage}
	\begin{minipage}{0.185\textwidth}
		\includegraphics[width=\textwidth]{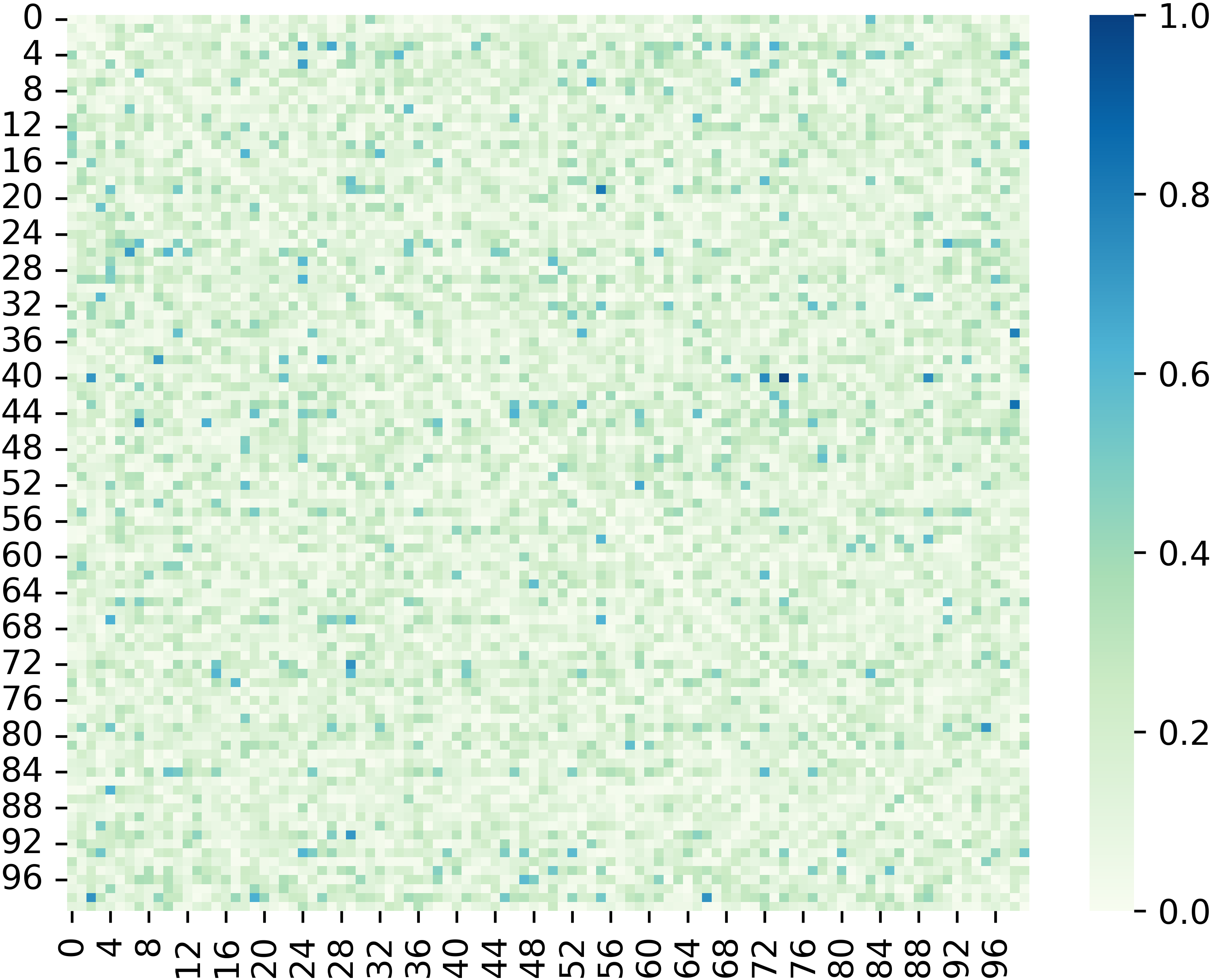}
		\caption*{(c) DKD \cite{dkd}}
	\end{minipage}
	\begin{minipage}{0.185\textwidth}
		\includegraphics[width=\textwidth]{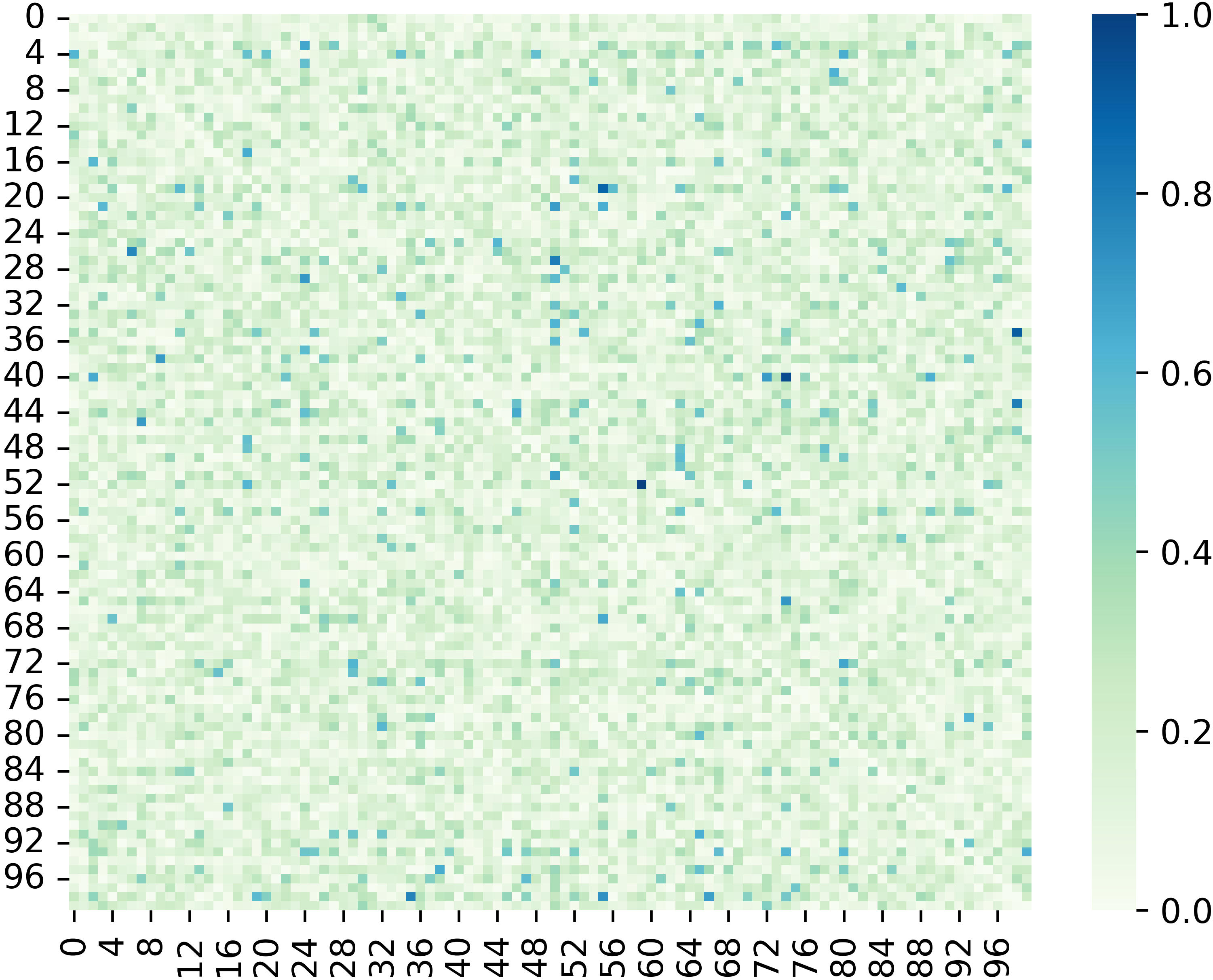}
		\caption*{(d) DOT \cite{dot}}
	\end{minipage}
	\begin{minipage}{0.185\textwidth}
		\includegraphics[width=\textwidth]{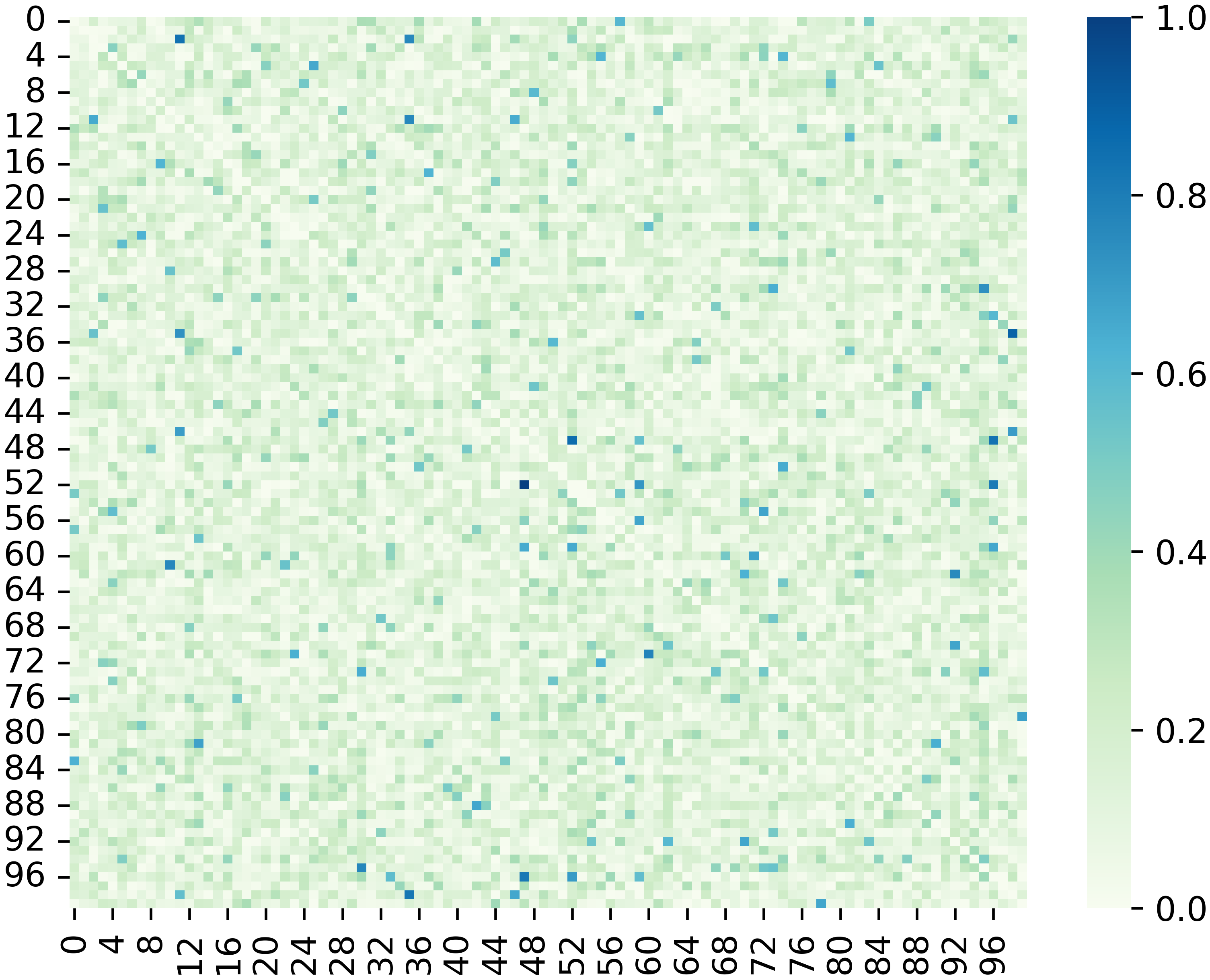}
		\caption*{(e) \textbf{TopKD}}
	\end{minipage}
	\caption{Differences between the student's and teacher's logit correlation matrices in a \textbf{heterogeneous structure}. Lighter shades indicate smaller differences.}
	\label{fig:correlation_comparison}
\end{figure}

\begin{table}[htbp]
	\centering
	\caption{Transferability of representations from CIFAR-100 \cite{cifar100} to STL-10 \cite{stl-10} and Tiny-ImageNet \cite{tiny-imagenet}. We compare methods in both \textbf{homogeneous} and \textbf{heterogeneous} structures.}
	\label{tab:transfer-merged}
	\resizebox{\textwidth}{!}{%
		\begin{tabular}{@{}c|ccccc|ccccc@{}}
			\toprule
			\multirow{2}{*}{Task} & \multicolumn{5}{c|}{Homogeneous (ResNet32x4 $\rightarrow$ ResNet8x4)} & \multicolumn{5}{c}{Heterogeneous (ResNet50 $\rightarrow$ MobileNetV2)} \\
			\cmidrule(lr){2-6} \cmidrule(lr){7-11}
			& Stu. & DOT \cite{dot} & FCFD \cite{fcfd} & \textbf{TopKD} & Tch. & Stu. & DKD+LS \cite{lskd} & WTTM \cite{ttm} & \textbf{TopKD} & Tch. \\
			\midrule
			STL-10      & 65.18 & 68.88 & \underline{71.26} & \textbf{71.38} & 66.59 & 65.61 & \underline{67.12} & 66.01 & \textbf{68.86} & 72.81 \\
			Tiny-IN & 34.75 & 33.54 & \textbf{36.89} & \underline{36.66} & 29.65 & 28.08 & \underline{33.57} & 31.45 & \textbf{35.35} & 40.84 \\
			\bottomrule
		\end{tabular}%
	}
\end{table}
\paragraph{Visualization.} We provide visualization results from two perspectives on CIFAR-100 \cite{cifar100} to further illustrate the effectiveness of TopKD. These visualizations assess the feature discriminability and the preservation of class-wise structural relationships in the student model. First, the t-SNE \cite{tsne} plots in Figure~\ref{fig:tsne_comparison}, using ResNet32x4 as the teacher and ResNet8x4 as the student, reveal that the feature representations learned with TopKD are more compact and class-discriminative compared to those produced by other distillation methods. Second, we visualize the normalized absolute difference between the logits correlation matrices of the teacher and student in Figure~\ref{fig:correlation_comparison}, where WRN-40-2 is used as the teacher and ResNet8x4 as the student. The smaller differences indicate that TopKD facilitates better structural alignment between teacher and student outputs.\par
\paragraph{Bigger models are not always better teachers.} Although stronger teacher models are expected to offer better supervision, they often struggle to transfer knowledge effectively to smaller students and can even underperform compared to weaker teachers. Prior work attributes this issue to the large capacity gap between teacher and student models \cite{efficacy, lfvi, dkd, lskd}. We argue most existing methods rely on directly matching the teacher’s logits, which overlooks their rich semantic structure and leads to inefficient knowledge transfer. Since teacher logits contain high-level class relationships, directly mimicking them can hinder the student’s ability to extract meaningful information. Our proposed TopKD alleviates this limitation by focusing on the informative structure within the Top-K knowledge, as demonstrated by the results in Table~\ref{tab:stronger-teacher-merged}.\par
\begin{table}[htbp]
	\centering
	\caption{Performance on CIFAR-100 \cite{cifar100} using different teacher-student configurations. WRN-16-2 is used as the student in both \textbf{homogeneous} (WRN-series teachers) and \textbf{heterogeneous} (other teacher architectures) settings.}
	\label{tab:stronger-teacher-merged}
	\resizebox{\textwidth}{!}{%
		\begin{tabular}{@{}l|cccc|cccc@{}}
			\toprule
			\multirow{2}{*}{Method} & \multicolumn{4}{c|}{Homogeneous (WRN-16-2 as Student)} & \multicolumn{4}{c}{Heterogeneous (WRN-16-2 as Student)} \\
			\cmidrule(lr){2-5} \cmidrule(lr){6-9}
			& WRN-28-2 & WRN-40-2 & WRN-16-4 & WRN-28-4 & VGG13 & WRN-16-4 & ResNet50 & ResNet32x4 \\
			\midrule
			Teacher Acc & 75.45 & 75.61 & 77.51 & 78.60 & 74.64 & 77.51 & 79.34 & 79.42 \\
			\midrule
			DKD \cite{dkd} & 75.92 & 76.24 & 76.00 & 76.45 & 75.45 & 76.00 & \underline{76.60} & 75.70 \\
			FCFD \cite{fcfd} & 74.80 & \textbf{76.34} & 75.99 & \underline{76.56} & n/a & 75.99 & n/a & \underline{76.73} \\
			WTTM \cite{ttm} & \underline{76.08} & 76.29 & \textbf{76.64} & 76.05 & \underline{75.63} & \textbf{76.64} & 76.17 & 76.04 \\
			\textbf{TopKD} & \textbf{76.24} & \underline{76.32} & \underline{76.36} & \textbf{77.62} & \textbf{75.73} & \underline{76.36} & \textbf{76.64} & \textbf{77.47} \\
			\bottomrule
		\end{tabular}%
	}
\end{table}
Notably, while other methods show fluctuating or even degraded student performance with stronger teachers, TopKD demonstrates a consistent improvement according to teacher capacities. This observation suggests that our method is more sensitive to teacher capacity and can benefit more reliably from stronger teachers.\par
\paragraph{Vision Transformers and Detection.} To further validate the robustness and generality of our TopKD, we conduct additional experiments on a wide spectrum of Vision Transformer architectures \cite{vit-at,lg,autokd,tinyvit,deti,t2t,pit,pvt}, and also extend our evaluations to object detection tasks using the MS-COCO dataset \cite{coco}. TopKD consistently delivers strong performance across diverse transformer-based backbones and tasks, further demonstrating its broad applicability and effectiveness (see appendix for details).\par
\section{Conclusion}
We introduce \textbf{Top-scaled Knowledge Distillation (TopKD)}, a novel and simple framework that enhances knowledge transfer by selectively focusing on the most informative outputs of the teacher model, its Top-K logits. Rather than treating all class predictions equally, TopKD explicitly identifies and amplifies high-confidence predictions that encode rich structural and semantic cues. It consists of two lightweight yet complementary modules: the Top-K Scaling Module (TSM), which reinforces the teacher’s signal by scaling the Top-K responses, and the Top-K Decoupled Loss (TDL), by decoupling the cosine similarity calculation of logits, the loss weight of Top-K knowledge is strengthened to achieve further emphasize its importance. Extensive experiments across diverse datasets and architectures show that TopKD performs favorably compared to most logit-based and feature-based distillation methods, and its modular design allows seamless integration into most existing distillation methods, demonstrates the effectiveness of Top-K knowledge.
%
%
\newpage
\medskip
{
\small
\bibliographystyle{abbrv}
\bibliography{neurips_2025}
}

\end{document}